\documentclass[journal]{IEEEtran}
\usepackage{cite}
\usepackage{amsmath,amssymb,amsfonts}
\usepackage{algorithm}
\usepackage{algpseudocode}
\usepackage{graphicx}
\usepackage{subcaption}
\usepackage{textcomp}
\usepackage{xcolor}
\usepackage{booktabs}
\usepackage{array}
\usepackage[justification=centering]{caption}
\usepackage{hyperref}
\hypersetup{
    colorlinks=true,
    linkcolor=blue,
    citecolor=blue,
    urlcolor=blue
}
\def\BibTeX{{\rm B\kern-.05em{\sc i\kern-.025em b}\kern-.08em
    T\kern-.1667em\lower.7ex\hbox{E}\kern-.125emX}}
\begin{document}

\title{CBE: A Complexity-adaptive Boundary-aware Ensemble Method for Imbalanced Data \\
}
\author{Xun Wei \thanks{Xun Wei is with the School of Industrial Software, Jiangxi University of Science and Technology, Nanchang 330044, China (email:xun.wei@jxust.edu.cn).}}

\maketitle

\begin{abstract}
In many real-world classification tasks, datasets exhibit class imbalance, which can degrade the generalization performance of conventional learning algorithms. Data complexity has recently been recognized as a more objective indicator of learning difficulty than the imbalance ratio alone, because class overlap and complex decision boundaries often impair classifiers more severely than imbalance itself. Samples near the decision boundary are inherently difficult to classify, which has motivated many researchers to focus on such instances to improve classifier performance. However, most under-sampling approaches fail to differentiate their sampling of boundary samples across datasets of varying complexity. To overcome this limitation, we propose a complexity-adaptive boundary-aware ensemble (\texttt{CBE}) framework that links the sampling weight distribution to data complexity and dynamically adjusts the boundary preference of under-sampling accordingly. We also introduce an overall complexity measure that combines metrics at multiple levels and use it to divide imbalanced datasets into three categories. Extensive experiments on 24 benchmark datasets from UCI and KEEL show that the proposed method outperforms many existing methods in terms of F-measure and G-mean on Easy and Median datasets, without degrading performance on hard cases. \footnote{\href{https://github.com/weixun2018/CBE}{https://github.com/weixun2018/CBE}}
\end{abstract}


\section{Introduction}
Class imbalance is a pervasive issue in real-world classification tasks, particularly in domains such as anomaly detection and disease analytics \cite{b26, b24}. This phenomenon arises when the majority class dominates the dataset, leading standard classifiers to prioritize frequent patterns over minority instances. Under such an imbalanced distribution, learning algorithms often degenerate: they tend to classify every instance as the majority class to minimize the error rate, thereby misclassifying most minority samples.

Sampling methods are typically adopted to resolve this issue by adjusting the sizes of the majority and minority classes to balance the training set \cite{b26}. These methods can be divided into under-sampling and over-sampling. Under-sampling handles class imbalance by discarding some majority examples to create a balanced training set, whereas over-sampling balances the dataset by adding new minority examples. Both approaches can be implemented in either a random or a heuristic manner. However, under-sampling may discard a large amount of useful data, while over-sampling may introduce extra bias, especially when the model consists of a single classifier.

Ensemble learning paradigms, in contrast to individual models, generally perform better by leveraging collective decision-making across multiple base models. This advantage stems from their ability to reduce variance through model diversity and mitigate overfitting by aggregating predictions. Two prominent ensemble frameworks are Bagging and Boosting \cite{b11}. In recent years, researchers have developed various strategies to apply these ensemble methods to class imbalance \cite{b2}.

More recently, data complexity has been regarded as a more objective indicator of learning difficulty than the imbalance ratio alone: class overlap and noise can be more damaging than imbalance itself \cite{barella2021assessing, santos2023unifying}. A growing body of work therefore designs imbalanced classification methods directly guided by data complexity \cite{liu2020selfpaced,sun2026forbu,park2026cihr}.

Samples near the decision boundary are inherently difficult to classify because of their ambiguous feature representations \cite{b26}. Conventional undersampling algorithms, which uniformly undersample the majority class at random, fail to prioritize these hard-to-classify instances, thereby undermining the discriminative power of the induced models. Distance-weighted sampling is an effective strategy for addressing this challenge in deep embedding learning \cite{b27}. However, overly aggressive boundary sampling is not always beneficial. In high-complexity datasets with severe class overlap, the boundary region no longer corresponds to a sharp, separable decision surface but to a wide, ambiguous overlap band dominated by irreducible noise \cite{santos2023unifying}. Concentrating sampling and weighting on these boundary samples then forces the classifier to fit the noise. To tackle this limitation, we propose a novel under-sampling strategy, complexity-adaptive distance-weighted sampling (CDWS), which adjusts the sampling strength of boundary samples according to data complexity.

In this paper, we propose a complexity-adaptive boundary-aware ensemble framework named \texttt{CBE} to overcome the aforementioned shortcomings. For clarity, we examine only binary classification problems with ensemble classifiers. The proposed method achieves higher performance than many existing ensemble methods for class imbalance.

The main contributions of this paper can be summarized as follows:
\begin{enumerate}
  \item We propose an overall complexity measure (OCM) for imbalanced data that combines metrics at multiple levels. Based on this measure, imbalanced datasets can be divided into three categories. Correlation experiments show that OCM is more effective than individual complexity measures.
  \item We introduce a novel distance-weighted sampling method that adaptively tunes the sampling intensity over boundary regions according to the intrinsic data complexity, so that boundary densification is emphasized on well-separated boundaries while relaxed under severe overlap.
  \item We propose a complexity-adaptive boundary-aware ensemble framework that integrates preference-based undersampling with ensemble learning to address imbalanced tasks.
\end{enumerate}

The remainder of this paper is organized as follows. Section \ref{sec:related_work} reviews related methods. Section \ref{sec:complexity} illustrates the complexity of imbalanced data. Section \ref{sec:method} presents our proposed method. Section \ref{sec:experiment} reports the experiments. Finally, Section \ref{sec:conclusion} concludes this paper.

\section{Related Work}\label{sec:related_work}

Sampling methods are often used to tackle class imbalance. Under-sampling balances the training set by reducing the number of majority instances, while over-sampling increases the number of minority instances. Both under-sampling and over-sampling can be random or heuristic. Random under-sampling selects a smaller subset from the majority class at random, while random over-sampling repeats minority instances at random to match the size of the majority class. NearMiss \cite{b3} is a heuristic under-sampling method that selects majority instances based on a distance metric, retaining those closest to the minority instances. NearMiss is essentially a form of hard negative mining. The one-sided selection procedure \cite{b4} tries to find a representative subset of majority examples by removing only "borderline" and "noisy" majority examples. SMOTE \cite{b5} adds new synthetic minority examples by randomly interpolating pairs of closest neighbors in the minority class. Under-sampling can weaken classifier performance by discarding many majority samples, whereas over-sampling can cause overfitting due to repeated or synthetic minority instances \cite{b9}. In addition, under-sampling methods generally run faster than over-sampling methods \cite{b2}.

Ensemble learning algorithms are also employed to address class imbalance owing to their strong performance. Bagging and Boosting are the main ensemble learning frameworks \cite{b11}. BalanceBagging \cite{b7} is a bagging ensemble method with additional balancing that bootstraps equal numbers of majority and minority instances to create training subsets. Balanced Random Forests \cite{b8} is an efficient ensemble algorithm that addresses class imbalance by modifying the random forest: it creates balanced training subsets by random under-sampling of the bootstrap examples. RUSBoost \cite{b6} alleviates class imbalance by randomly under-sampling majority instances at each iteration of the AdaBoost algorithm. EasyEnsemble \cite{b2} further exploits the majority examples by randomly under-sampling them to build multiple balanced subsets. EasyEnsemble is similar to Balanced Random Forests, except that the former uses the samples to generate boosted ensembles, whereas the latter uses them to train decision trees. REA \cite{b12} is a bagging method that addresses class imbalance by under-sampling the majority class and over-sampling the minority class to form balanced training subsets, and it uses multiple different machine learning methods to train these subsets. Adaptive EUSBoost \cite{b16} is similar to EasyEnsemble, but it introduces a weighting algorithm that assigns each individual classifier a voting weight according to its own training error. This strategy yields better performance, especially in highly imbalanced settings, and it also uses the OTSU algorithm to determine the optimal threshold. ClusterBal \cite{b15} is a data balancing method that clusters the majority samples into several subsets to create balanced training subsets. Tang et al. proposed a novel metric GIR that measures class distribution imbalance, together with two adaptive sampling-based ensemble approaches, GIREnUs and GIREnOs, which combine boosting and bagging strategies \cite{b17}.

In recent years, a growing line of research in imbalanced learning has developed sampling and ensemble algorithms that explicitly exploit the characteristics of the decision boundary rather than merely rebalancing the global class distribution. Cerqueira et al.~\cite{cerqueira2022layered} proposed ICLL, a layered learning framework that first uses hierarchical clustering to separate hard boundary samples from easy majority samples and then trains the final classifier only on boundary samples together with all minority samples, showing both theoretically and empirically that performance loss arises mainly from distribution shift in the boundary region rather than from the global imbalance ratio, and that conventional under-/over-sampling respectively distorts the boundary by deleting critical samples and amplifies boundary noise. Dai et al.~\cite{dai2022multigranularity} extended Tomek-Link detection with the multi-granularity relabeled under-sampling algorithm MGRU, which jointly identifies global hard overlaps and locally latent soft overlaps and removes only ambiguous majority samples in the boundary region, demonstrating that multi-granularity detection covers the majority of boundary overlaps that single-granularity methods miss. From an uncertainty perspective, Grina et al.~\cite{grina2023belief} combined belief function theory with ensemble learning for multi-class imbalance, using belief soft labels to quantify boundary ambiguity, preferentially under-sampling highly ambiguous majority samples and directionally over-sampling boundary minority samples, and argued that boundary ambiguity rather than global imbalance is the key factor governing the severity of the problem.

Another line of research directly guides imbalanced classification with data complexity measures, which capture learning difficulty more objectively than the imbalance ratio. Barella et al. \cite{barella2021assessing} adapted classical complexity measures to imbalanced scenarios by decomposing them per class, and experimentally showed that the reduction in data complexity after imbalance treatments correlates with the achieved performance improvement, providing a theoretical basis for complexity-guided sampling. Liu et al. \cite{liu2020selfpaced} proposed the Self-paced Ensemble (SPE), which characterizes the intrinsic difficulty of each instance through its classification hardness, a quantity implicitly reflecting class overlap and noise, and then harmonizes the data hardness distribution by progressively under-sampling the majority class in a self-paced manner. Sun et al. \cite{sun2026forbu} proposed FO-RBU, which estimates the overall class overlap degree from the distribution of overlap ratios on each feature, and uses it to determine the proper under-sampling extent of the Radial-Based Undersampling approach, thereby replacing the imbalance ratio as the sampling-amount criterion. Park et al. \cite{park2026cihr} introduced CIHR, a hybrid resampling method in which a relative class-impurity measure quantifies the overlap degree of individual instances: majority instances with high impurity are iteratively removed, while minority instances with high impurity are prioritized as seeds for synthetic oversampling. These studies collectively validate that adaptively modulating the sampling intensity according to data complexity, the same principle underlying our approach, is an effective and well-grounded strategy for imbalanced classification.

\section{Complexity of Imbalanced Data}\label{sec:complexity}

\subsection{Class Boundary and Class Overlapping}

In classification tasks, a class boundary denotes the geometric partition (a hyperplane in a high-dimensional feature space) that separates feature regions assigned to different class labels by a classifier, characterizing the intrinsic separable structure of the data distribution. It describes where the decision rule switches its prediction from one category to another. A class boundary should be clearly distinguished from class overlap: the former refers to the dividing manifold between classes, whereas the latter describes ambiguous feature regions where samples with different labels coexist densely. A clear, simple class boundary corresponds to low data complexity and limited intermingling of cross-class samples. In contrast, severe class overlap often yields twisted, fragmented, and fuzzy boundaries, which are a primary cause of classification errors, especially in class-imbalanced scenarios \cite{santos2023unifying}. Many existing sampling and ensemble algorithms explicitly target boundary regions, because boundary samples carry decisive information for constructing reliable decision surfaces, rather than merely addressing overlapping samples.

\subsection{Complexity Measures of Imbalanced Data}\label{sec:OCM}

Measures of classification complexity can be used to estimate how difficult it is to separate the data points into their expected classes. Descriptors of the spatial distribution of the data and estimates of the shape and size of the decision boundary are among the known measures for this purpose \cite{b29}.
Researchers have adapted complexity measures to the context of imbalanced datasets and demonstrated their effectiveness \cite{barella2021assessing}. Lower data complexity implies clearer class boundaries; conversely, higher complexity corresponds to fuzzier boundaries and more severe class overlap.

This paper adopts three complexity measures at different levels to assess the data complexity of imbalanced datasets, namely $N1$, $N3$, and $L2$. The details of these measures are given below.

\begin{enumerate}
\item \emph{N3(sample level)}: Leave-one-out error rate of the NN classifier.

N3 is the ratio of the number of examples whose 1-NN belongs to a different class to the total number of examples. The adapted N3 focuses on the minority class. This metric quantifies local class overlap in individual sample neighborhoods. It bounds the intrinsic classification ambiguity and shows the strongest correlation with the performance improvement of sampling strategies.
\item \emph{N1(boundary level)}: Fraction of points on the class boundary.

N1 builds a minimum spanning tree (MST) that connects all instances in a dataset based on their pairwise distances, regardless of their classes. It then counts the number of instances connected to at least one instance from another class. The adapted N1 focuses on the minority class. This metric measures the scale and tortuosity of the decision boundary and governs the convergence rate and computational cost of the learner when fitting the boundary geometry.
\item \emph{L2(global level)}: Training error of a linear classifier.

L2 is the training error of a linear classifier. To compute it, we induce a linear classifier from the training set and use its classification error rate. The higher the value, the less linear the classification boundary. The adapted L2 uses the error of the minority class. This metric reflects the global linear separability of the data distribution and provides a global perspective for identifying special cases, such as extreme tail minority classes in ordered discrete feature spaces.
\end{enumerate}

We then calculate the weighted average of these three measures as the overall complexity measure (OCM), as shown in Equation~\eqref{equ:cm}.
\begin{equation}\label{equ:cm}
  OCM =  \alpha*N3 + \beta*N1 + \gamma*L2
\end{equation}
Intuitively, we set $\alpha=0.5$, $\beta=0.3$, and $\gamma=0.2$ according to the contribution of each measure to data complexity.

We divide the datasets into three categories according to their OCM values, as shown in Table~\ref{tab:dataset_cat}.
Easy datasets correspond to low N1, N3, and L2, with simple and clear class boundaries and negligible local overlap.
Median datasets usually have low N1 and N3 with high L2, or median N1, N3, and L2, representing data with geometrically complex decision boundaries but limited inter-class mixing.
Hard datasets have high N1, N3, and L2, where severe class overlap leads to high intrinsic classification ambiguity and the performance gains brought by sampling strategies are generally limited.

The relationship \(N1 \geq N3\) holds mathematically: samples misclassified in leave-one-out 1-NN (counted in N3) are always adjacent to opposite-class instances in the minimum spanning tree and thus included in N1. However, the two metrics capture distinct dimensions of classification difficulty and are not redundant. N3 quantifies the severity of local class overlap and intrinsic classification ambiguity at the sample level, while N1 measures the overall scale and geometric complexity of decision boundaries.
Notably, datasets with high adapted N1 and N3 rarely yield low adapted L2. High adapted N3 implies severe local overlap around minority samples. When training a linear SVM, such overlap inevitably increases the minority error rate measured by the adapted L2. Hence, the high-N3, low-L2 combination scarcely appears in our experiments, which justifies the three-way categorization.

\begin{table}[htbp]
  \centering
  \caption{Dataset Categorization via Complexity Measures}
  \label{tab:dataset_cat}
  \begin{tabular}{ccccc}
    \toprule
    Category & $N1$  & $N3$  & $L2$  &OCM  \\
    \midrule
    Easy     & Low   & Low   & Low  & Low   \\
    Median   & Low/Median   & Low/Median   & High/Median & Median  \\
    Hard     & High  & High  & High & High  \\
    \bottomrule
  \end{tabular}
\end{table}

\begin{figure}[htbp]
  \centering
  \includegraphics[width=\linewidth]{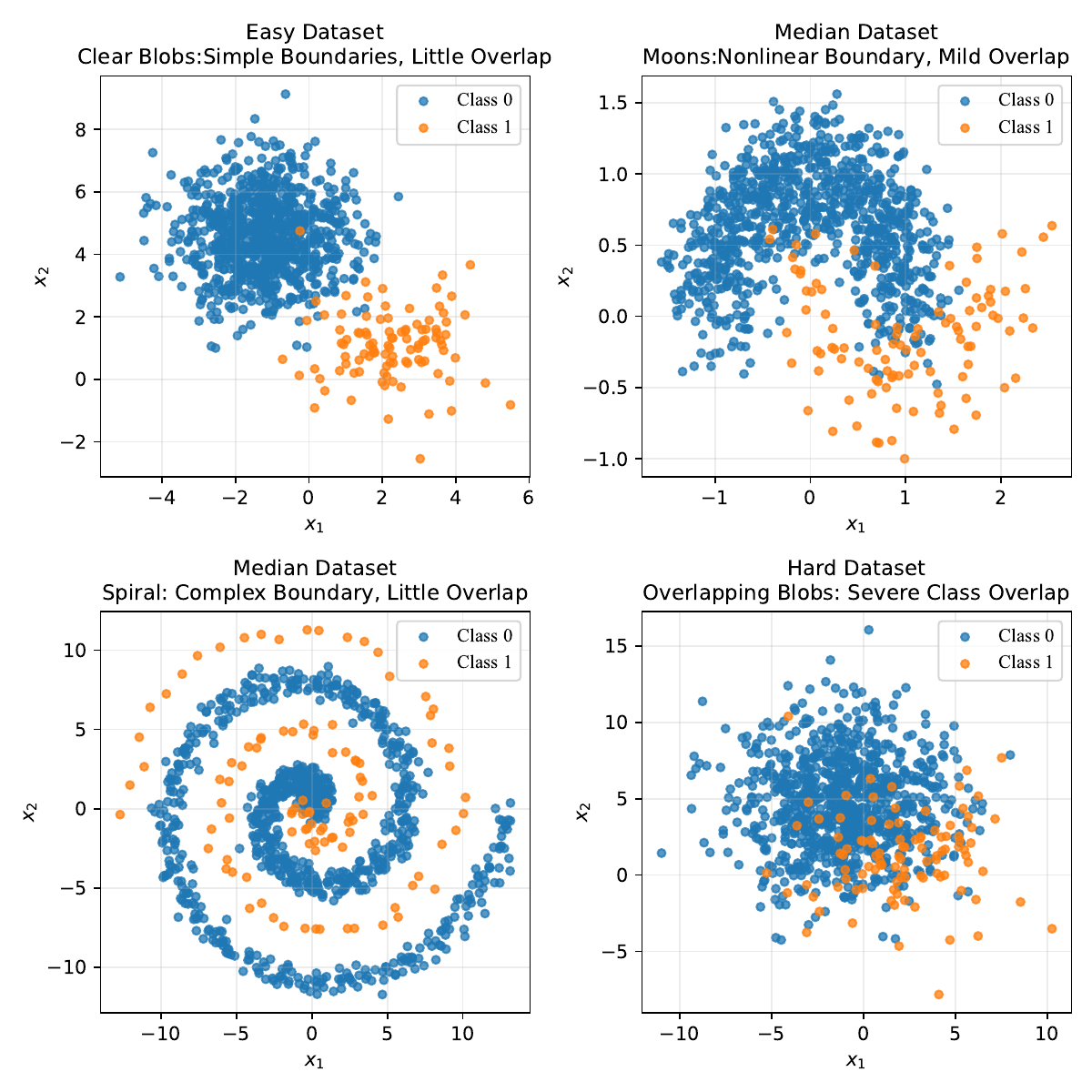}
  \caption{Synthetic imbalanced datasets illustrating the proposed categorization based on complexity measures. The four datasets range from Easy to Hard, each containing 900 negative samples and 100 positive samples. In particular, we show two kinds of Median datasets: one with low N1 and N3 and high L2, and the other with median N1, N3, and L2.}
  \label{fig:dataset_cat}
\end{figure}

We synthesize four datasets to illustrate this categorization, as shown in Fig.~\ref{fig:dataset_cat}.
The detailed values of the complexity measures on these four synthetic datasets are listed in Table~\ref{tab:dataset_syn}. The difficulty category is not determined by a single fixed OCM threshold. Here, we use 0.2 and 0.6 to divide these four synthetic datasets. In general, for a group of datasets with varied complexity, we adopt the 1/3 and 2/3 quantiles as the split points.

\begin{table}[htbp]
  \centering
  \caption{Complexity Measures on Four Synthetic Datasets.}
  \label{tab:dataset_syn}
  \begin{tabular}{cccccc}
    \toprule
    Dataset      & $N1$  & $N3$  & $L2$  & OCM   & Category\\
    \midrule
    Clear Blob   & 0.1    & 0.1    & 0.06 & 0.09 & Easy  \\
    Moon         & 0.31   & 0.23   & 0.38 & 0.28 & Median\\
    Spiral       & 0.2    & 0.11   & 1.0  & 0.32 & Median\\
    Overlap Blob & 0.82   & 0.65   & 1.0  & 0.77 & Hard  \\
    \bottomrule
  \end{tabular}
\end{table}

\section{Method}\label{sec:method}

\subsection{Complexity-adaptive Distance-Weighted Sampling}

Distance-weighted sampling (DWS) is a sampling distribution that corrects bias while controlling variance, originally proposed in deep embedding learning \cite{b27}. Given an anchor example $a$, it samples negative examples according to distance, with weights $D(a)^{-1}$. DWS can improve model performance on Easy and Median datasets. On Hard datasets, however, it is ineffective or even harmful: it aggravates the inter-class mixing of examples and thereby exacerbates performance degradation.

To address the inability of DWS to adapt to datasets with different degrees of class overlap, we link the sampling weight distribution to data complexity and dynamically adjust the boundary preference of under-sampling according to the classification difficulty of the dataset. The overall complexity measure (OCM) introduced above is used to quantify the severity of class overlap and boundary ambiguity, which serves as the basis for adaptively adjusting the sampling intensity.

We introduce a temperature parameter $T$ into distance-weighted sampling to control how fast the sampling weight decays with the nearest distance from majority samples to minority samples. A smaller $T$ makes the sampling weights highly sensitive to distance: majority samples close to the class boundary receive much higher sampling probability, and the distribution is strongly concentrated on the boundary region. A larger $T$ flattens the weight distribution, so the sampling strategy gradually approaches uniform random sampling. The rationale for adaptive temperature adjustment is as follows. On Easy datasets, a small $T$ strengthens the sampling priority of boundary samples, enriching boundary information in training subsets and improving minority-class accuracy. On Hard datasets, a large $T$ weakens excessive boundary preference.
This prevents sampling from over-concentrating on noisy overlapping regions, avoids a systematic shift of the decision boundary, and preserves the classification specificity of the majority class together with the minority-class recall.

To make $T$ adapt smoothly and continuously to OCM and avoid performance jumps caused by hard threshold switching, we use a translated and normalized Sigmoid function to build the mapping from complexity to temperature. Because the standard Sigmoid function \(\sigma(\cdot)\) is inherently asymmetric, linear normalization is applied to restrict the output adaptation coefficient $k$ to the range \([0,1]\). The formulation is given in Equation~\eqref{equ:k}.

\begin{subequations}\label{equ:k}
\begin{align}
z &= s \cdot (\text{OCM} - x_0) \\
k &= \frac{\sigma(z) - \sigma(-s x_0)}{\sigma(s(1-x_0)) - \sigma(-s x_0)}
\end{align}
\end{subequations}
where \(x_0=0.4\) is the inflection point that determines when the strategy transitions, and \(s=10\) controls the steepness of the adaptive curve. The coefficient $k$ is then used to compute the sampling temperature $T$ within a statistically determined interval, as shown in Equation~\eqref{equ:T}.

\begin{equation}\label{equ:T}
T = T_{\text{min}} + k \cdot (T_{\text{max}} - T_{\text{min}})
\end{equation}

Here, \(T_{\text{min}}\) and \(T_{\text{max}}\) are derived from the 10\% and 90\% quantiles of the nearest-neighbor distance distribution between majority and minority samples, providing data-driven and robust temperature boundaries. Based on the adaptive temperature, the distance-based sampling weight for each majority sample is calculated as in Equation~\eqref{equ:w}.
\begin{equation}\label{equ:w}
w_{N_i} = \exp\left(-\frac{d_{N_i}}{T}\right)
\end{equation}
where \(d_{N_i}\) is the nearest Euclidean distance from the $i$-th majority sample to the minority samples. To avoid extreme weight concentration and guarantee basic sampling opportunities for boundary samples, hard weight clipping is applied.


This sampling strategy avoids the drawbacks of fixed-intensity distance weighting. It maintains a strong boundary preference to improve minority recall on simple datasets, and gradually weakens the weighting to stabilize majority specificity on highly overlapping datasets. In this way, it achieves a stable trade-off between minority-class accuracy and overall balanced performance.

\subsection{Complexity-adaptive Boundary-Aware Ensemble}

Under-sampling is an effective way to handle imbalanced data, but it discards a large amount of potentially useful data. In addition, selecting only a small fraction of majority examples at a time yields unstable results, that is, high variance. EasyEnsemble \cite{b2} is a typical ensemble model for imbalance learning that addresses this problem by repeatedly under-sampling majority instances at random. It adopts a "bagging of boosting" strategy, using AdaBoost as the base learner to strengthen the fitting ability. However, EasyEnsemble randomly undersamples the majority examples and fails to prioritize examples near the decision boundary. It also fails to adapt to datasets with different degrees of complexity. To overcome these limitations, we incorporate the proposed CDWS method into a complexity-adaptive boundary-aware ensemble framework, \texttt{CBE}. The pseudocode is shown in Algorithm \ref{alg1}.

\begin{algorithm}
  \caption{\emph{CBE}}\label{alg1}
  \begin{algorithmic}[1]
    \Procedure {CBE}{$\mathcal{P}$, $\mathcal{N}$, $n$, $t$}
    \State \textbf{Input:}
    \State $\mathcal{P}$: the set of minority class
    \State $\mathcal{N}$: the set of majority class
    \State $n$: the number of subsets
    \State $t$: iterations to train a AdaBoost model $\mathcal{H}_i$
    \State \textbf{Steps:}
    \State $OCM \leftarrow$ calculate the overall complexity measures of dataset
    \State $d_N \leftarrow$ calculate the minimum euclidean distance to $\mathcal{P}$ for every sample in $\mathcal{N}$
    \State $T \leftarrow$ derive the sampling temperature via $OCM$ and $d$
    \State $w_N = \exp\left(-\frac{d_N}{T}\right) \leftarrow$ calculate the sampling weights
    \For    {$i\ =\ 1\ to\ n$}
    \State $\mathcal{N}_i \leftarrow$ undersample a subset from $\mathcal{N}$ with weights $w_N$, and $|\mathcal{N}_i|=|\mathcal{P}|$
    \State $\mathcal{D}_i \leftarrow \mathcal{P} \bigcup\mathcal{N}_i$
    \State $\mathcal{H}_i\leftarrow$ train AdaBoost on $\mathcal{D}_i$ under $t$ iterations
    \EndFor
    \State \textbf{Output:}
    \State $H$:an ensemble model
    \State $H(x)=sgn(\sum_{i=1}^n \mathcal{H}_i(x)-\theta) $
    \EndProcedure
  \end{algorithmic}
\end{algorithm}

\begin{figure*}[htbp]
  \centering
  \subfloat[CDWS on Four Synthetic Datasets.]
  {
    \includegraphics[width=0.32\textwidth]{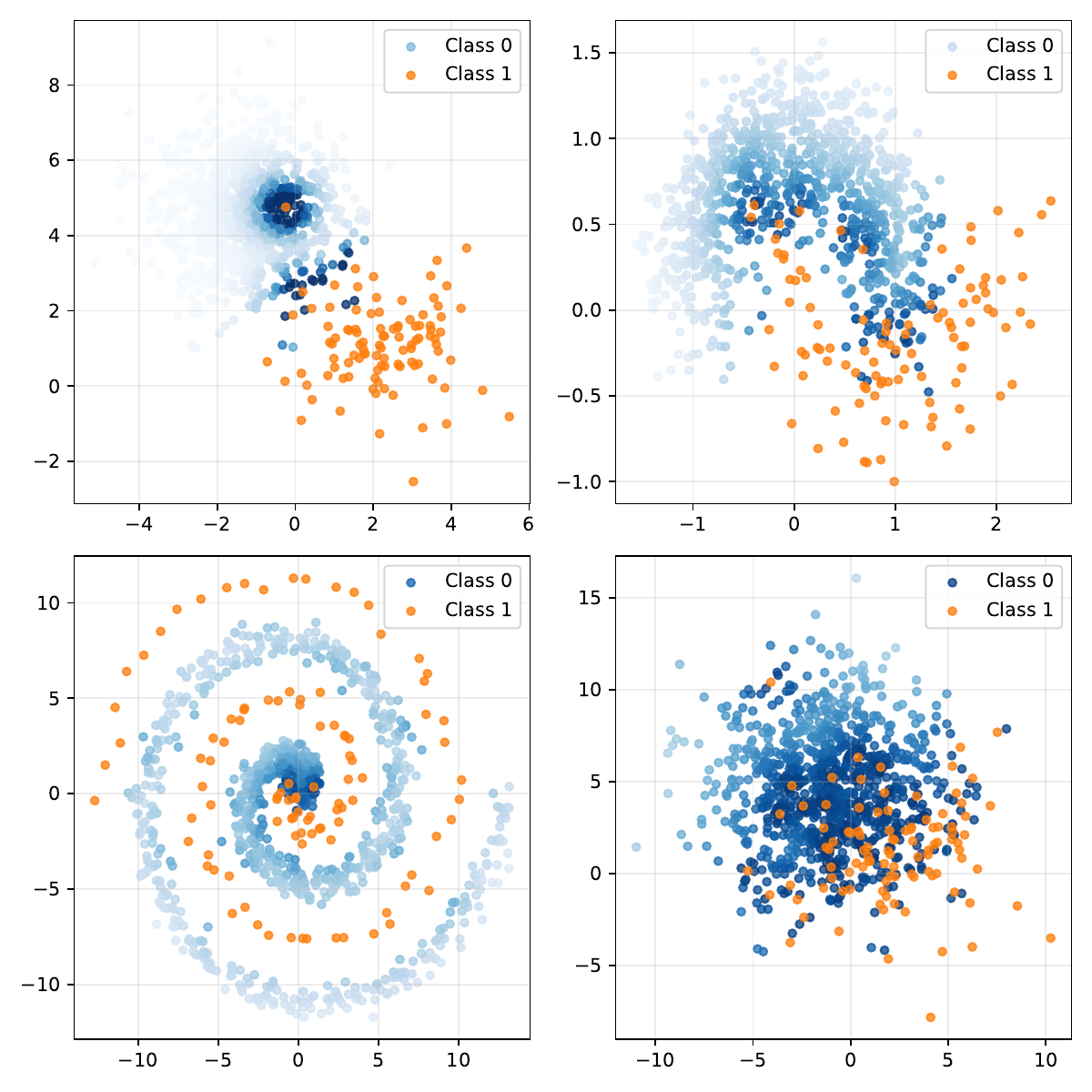}
    \label{fig:dws}
  }
  \subfloat[Overview of CBE framework.]
  {
    \includegraphics[width=0.68\textwidth]{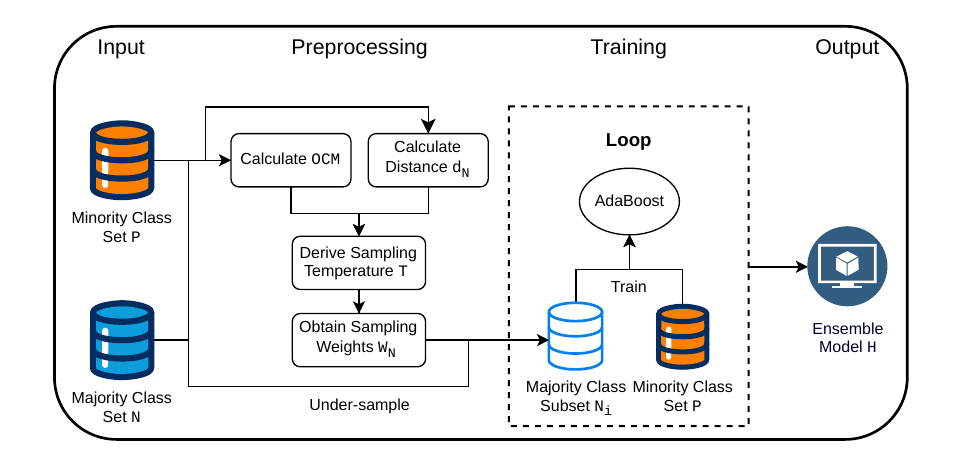}
    \label{fig:enhanced}
  }
  \caption{(a) shows the sampling probabilities of majority instances, encoded by the intensity of the blue color. CDWS applies an adaptive sampling intensity to boundary majority instances. On datasets with low complexity, it assigns closer examples higher weights to preserve critical boundary information, while it weakens excessive boundary preference on hard datasets. \\
  (b) shows the procedure of the CBE framework. It first calculates the overall complexity of the dataset; then performs weighted under-sampling of the majority examples based on this complexity to create a balanced training subset; finally, it trains base classifiers on multiple subsets and obtains the ensemble model.}
  \label{fig:dws-enhanced}
\end{figure*}

The parameter $\theta$ in Algorithm \ref{alg1} is the ensemble threshold above which an example is classified as positive. CART \cite{breiman1984classification} is usually used as the base model of AdaBoost. An example of the CDWS method and an overview of the \texttt{CBE} framework are shown in Fig.~\ref{fig:dws-enhanced}.

\section{Experiments} \label{sec:experiment}

\subsection{Datasets} \label{sec:datasets}
We conduct experiments on 24 imbalanced datasets from UCI \cite{b23} and KEEL. Each difficulty category contains 8 datasets, split by the 1/3 and 2/3 quantiles of OCM, namely 0.220 and 0.581.
The details of these datasets are shown in Table \ref{tab:datasets}, including basic information and complexity measures. The basic information covers dataset size, number of features, and imbalance ratio. The dataset size ranges from hundreds to thousands, the number of features from several to hundreds, and the imbalance ratio from mild to severe across the three categories. The datasets are listed in ascending order of OCM, from 0.025 to 0.871.

\begin{table*}
  \caption{Details of the datasets, including basic information and data complexity measures. The 24 datasets are divided into three groups by OCM (Easy, Median, Hard) and listed in ascending order of OCM.}
  \label{tab:datasets}
  \begin{center}
    \begin{tabular}{c|ccc|ccccc}
    \hline
     Dataset        &Size         &Features        &Imbalance Ratio   & N1      & N3        & L2         & OCM  \\
    \hline
     mfeat-fac      &2000         &216             &9.00              &0.050    &0.020      &0.000       &0.025 \\   
     mfeat-pix      &2000         &240             &9.00              &0.045    &0.025      &0.000       &0.026 \\   
     mfeat-kar      &2000         & 64             &9.00              &0.070    &0.050      &0.010       &0.048 \\   
   newthyroid2      & 215         &  5             &5.14              &0.171    &0.029      &0.029       &0.072 \\   
        vowel0      & 988         & 13             &9.98              &0.133    &0.000      &0.167       &0.073 \\   
       vehicle      & 845         & 18             &3.25              &0.176    &0.060      &0.035       &0.090 \\   
         musk2      &6598         &166             &5.49              &0.251    &0.161      &0.180       &0.192 \\   
        ecoli4      & 336         &  7             &15.80             &0.250    &0.250      &0.100       &0.220 \\
    \hline    
          wave      &5000         & 21             &2.04              &0.319    &0.213      &0.108       &0.224 \\   
        ecoli1      & 336         &  7             &3.36              &0.390    &0.286      &0.234       &0.307 \\   
        glass0      & 214         &  9             &2.06              &0.471    &0.186      &0.400       &0.314 \\   
        ecoli3      & 336         &  7             &8.60              &0.600    &0.429      &0.371       &0.469 \\   
        glass1      & 214         &  9             &1.82              &0.474    &0.329      &0.921       &0.491 \\   
          pima      & 768         &  8             &1.87              &0.616    &0.485      &0.433       &0.514 \\   
          bank      &7842         & 47             &3.39              &0.652    &0.508      &0.488       &0.547 \\   
      vehicle1      & 846         & 18             &2.90              &0.728    &0.548      &0.442       &0.581 \\
    \hline    
     mfeat-fou      &2000         & 76             &9.00              &0.715    &0.575      &0.420       &0.586 \\   
      vehicle3      & 846         & 18             &2.99              &0.717    &0.571      &0.453       &0.591 \\   
          wine      &6497         & 11             &5.02              &0.724    &0.366      &1.000       &0.600 \\   
        yeast1      &1484         &  8             &2.46              &0.683    &0.513      &0.760       &0.613 \\   
     mfeat-mor      &2000         &  6             &9.00              &0.690    &0.635      &1.000       &0.724 \\   
      haberman      & 306         &  3             &2.78              &0.852    &0.728      &0.975       &0.815 \\   
     mfeat-zer      &2000         & 47             &9.00              &0.880    &0.785      &1.000       &0.857 \\   
       abalone      &4177         & 10             &7.58              &0.912    &0.795      &1.000       &0.871 \\   
    \hline
    \end{tabular}
    \end{center}
\end{table*}

\subsection{Evaluation Criteria}\label{evaluation}
It is well known that the error rate is not an appropriate evaluation criterion when a dataset is imbalanced or misclassification costs are unequal.
In this paper, we use \textit{F-measure} and \textit{G-mean} as the performance metrics. Notably, we exclude \textit{AUC}, as clearly pointed out in many surveys on imbalanced learning, ROC-AUC is suitable for rough comparison of global ranking ability, yet it is not recommended as the primary evaluation metric for imbalanced classification \cite{branco2016survey}.

\begin{table}
  \caption{Confusion Matrix}
  \label{tab:confusion}
  \begin{center}
    {\fontsize{7}{9}\selectfont
    \begin{tabular}{|c|c|c|}
    \hline
    & Predicted Positive Class & Predicted Negative Class \\
    \hline
    Actual Positive Class & TP(true positive) & FN(false negative) \\
    \hline
    Actual Negative Class & FP(false positive) & TN(true negative) \\
    \hline
    \end{tabular}
    }
    \end{center}
\end{table}

\textit{F-measure} and \textit{G-mean} are functions of the confusion matrix as shown in Table~\ref{tab:confusion}.
\textit{F-measure} is the harmonic mean of precision and recall and is commonly used to estimate the overall performance of a model.
\textit{G-mean} is the square root of the product of the true positive rate (TPR) and the true negative rate (TNR).
These two metrics are defined in Equations \eqref{equ:fmeasure} and \eqref{equ:gmean}. The \textit{F-measure} focuses on minority examples, whereas the \textit{G-mean} considers both minority and majority examples.

\begin{subequations}\label{equ:metrics}
\begin{align}
  True Positive Rate(TPR)  &=\frac{TP}{TP+FN}\\
  True Negative Rate(TNR)  &=\frac{TN}{TN+FP}\\
  Precision   &=\frac{TP}{TP+FP}\\
  Recall      &=\frac{TP}{TP+FN}\\
  F-measure   &=\frac{2 \times Precision \times Recall}{Precision + Recall}\\ 
  G-mean      &=\sqrt{TPR \times TNR} \label{equ:gmean}\\
  \label{equ:fmeasure}
\end{align}
\end{subequations}

\subsection{Experimental Settings}\label{Experimental Settings}

For every dataset, we perform fivefold stratified cross-validation. To account for randomness, the entire cross-validation is repeated 10 times. The final \textit{AUC}, \textit{G-mean}, and \textit{F-measure} are the averages over these 10 runs. All algorithms are implemented in Python, using Scikit-Learn as the base framework and the Imbalanced-learn package \cite{b22}.

We compare CART with nine ensemble models. This paper focuses on ensemble methods combined with undersampling. The nine methods for class imbalance are listed below:
\begin{enumerate}
  \item \emph{CART}. Classification and regression trees\cite{breiman1984classification}. It uses the entire data set to train a single classifier.
  \item \emph{AdaBoost}(abbreviated as Ada). A traditional AdaBoost classifier. The base estimator is CART and estimator number is set 100.
  \item \emph{Random Forest}(abbreviated as RF). A traditional random forest classifier. The estimator number is set 100.
  \item \emph{Undersampling + AdaBoost}(abbreviated as UAda). First, a subset of the majority samples with the same size as the minority class is undersampled; an AdaBoost classifier is then trained on the resulting balanced training set. The base estimator configuration is the same as that of Ada.
  \item \emph{Undersampling + Random Forest}(abbreviated as URF). The difference from UAda is that RF is used to train the balanced training set. The base estimator configuration is the same as that of RF.
  \item \emph{BalancedBagging}(abbreviated as BB). A Bagging classifier with additional balancing. Here we use random under-sample to balance subset. The base estimator is CART and estimator number is set 100.
  \item \emph{BalancedRandomForests}(abbreviated as BRF). A balanced random forest by randomly under-sampling each bootstrap sample. The estimator number is set 100.
  \item \emph{RUSBoost}(abbreviated as RUSB). During learning, the problem of class balancing is alleviated by random under-sampling the sample at each iteration of the AdaBoost algorithm. The base estimator is CART and estimator number is set 100.
  \item \emph{EasyEnsemble}(abbreviated as EE). Like random under-sampling, this method samples the negative examples with replacement and repeats the process $M$ times to create $M$ balanced subsets. We set $M=10$. The base estimator is AdaBoost with CART as the base classifier, and the number of AdaBoost estimators is set to 10.
  \item \emph{CBE}. The complexity-adaptive boundary-aware ensemble model we proposed; details are given in Section \ref{sec:method}. Here, we also create 10 balanced subsets. The base classifier of AdaBoost is CART, and the number of AdaBoost estimators is set to 10.
\end{enumerate}

CART and Ada are the baseline methods. In CART, we set min\_samples\_split = 10 to prune the tree and avoid overfitting. In addition, for the ensemble methods based on CART, we set max\_features='sqrt' to reduce variance and training overhead \cite{b0}.

\subsection{Analysis of Training Time}
The running times of these methods are recorded in Table \ref{tab:time} on a Mac with an M4 CPU. With n\_job=1, the single-estimator CART runs fastest. Imbalance learning methods based on under-sampling run faster than conventional methods because they use fewer training examples. The RF series (RF, URF, BRF) runs slightly faster than the Ada series (Ada, UAda). RUSB runs faster than UAda, mainly because of the different convergence algorithms in the runtime environment: SAMME.R typically converges faster than SAMME. Owing to the extra computation of data complexity and sample distances, CBE runs slightly slower than EE.

\begin{table*}
  \caption{Training Times (in Seconds). The last row shows the average training time of each method over the 24 datasets.}
  \label{tab:time}
  \begin{center}
    \begin{tabular}{c|cccccccccc}
    \hline
                 &CART   &Ada    &RF     &UAda   &URF    &BB     &BRF    &RUSB   &EE     &CBE\\
    \hline
    mfeat-fac    &0.064  &0.540  &0.245  &0.036  &0.064  &0.169  &0.106  &0.016  &0.105  &0.169\\
    mfeat-pix    &0.012  &0.198  &0.091  &0.067  &0.045  &0.134  &0.076  &0.017  &0.077  &0.139\\
    mfeat-kar    &0.045  &0.596  &0.288  &0.104   &0.06  &0.139  &0.102  &0.028  &0.109  &0.138\\
    newthyroid2  &0.000  &0.003  &0.034  &0.001  &0.033  &0.081  &0.051  &0.004  &0.019  &0.026\\
    vowel0       &0.002  &0.066  &0.058  &0.010  &0.035  &0.089  &0.058  &0.007  &0.038  &0.050\\
    vehicle      &0.001  &0.094  &0.052  &0.059  &0.042  &0.099  &0.068  &0.010  &0.064  &0.071\\
    musk2        &0.285  &2.198  &1.099  &0.700  &0.328  &0.668  &0.522  &0.137  &0.775  &1.292\\
    ecoli4       &0.000  &0.018  &0.036  &0.004  &0.033  &0.083  &0.053  &0.005  &0.026  &0.031\\
    wave         &0.035  &0.847  &0.398  &0.542  &0.291  &0.501  &0.465  &0.207  &0.537  &0.699\\
    ecoli1       &0.000  &0.044  &0.036  &0.038  &0.034  &0.083  &0.054  &0.006  &0.047  &0.048\\
    glass0       &0.000  &0.044  &0.036  &0.040  &0.035  &0.084  &0.055  &0.012  &0.048  &0.049\\
    ecoli3       &0.000  &0.043  &0.036  &0.026  &0.033  &0.081  &0.052  &0.004  &0.043  &0.044\\
    glass1       &0.000  &0.044  &0.037  &0.041  &0.035  &0.084  &0.056  &0.014  &0.049  &0.050\\
    pima         &0.001  &0.072  &0.050  &0.059  &0.045  &0.101  &0.071  &0.034  &0.067  &0.072\\
    bank         &0.027  &0.527  &0.285  &0.241  &0.137  &0.303  &0.242  &0.121  &0.252  &0.659\\
    vehicle1     &0.002  &0.102  &0.064  &0.067  &0.048  &0.110  &0.077  &0.021  &0.075  &0.081\\
    mfeat-fou    &0.029  &0.587  &0.254  &0.110  &0.065  &0.155  &0.116  &0.016  &0.119  &0.143\\
    vehicle3     &0.002  &0.102  &0.066  &0.066  &0.048  &0.108  &0.077  &0.026  &0.075  &0.081\\
    wine         &0.016  &0.669  &0.354  &0.229  &0.135  &0.270  &0.231  &0.088  &0.237  &0.386\\
    yeast1       &0.002  &0.100  &0.067  &0.071  &0.052  &0.121  &0.086  &0.041  &0.083  &0.095\\
    mfeat-mor    &0.001  &0.098  &0.058  &0.050  &0.038  &0.099  &0.067  &0.005  &0.058  &0.068\\
    haberman     &0.000  &0.042  &0.036  &0.038  &0.034  &0.084  &0.055  &0.016  &0.048  &0.048\\
    mfeat-zer    &0.021  &0.413  &0.186  &0.087  &0.061  &0.141  &0.106  &0.012  &0.097  &0.116\\
    abalone      &0.008  &0.495  &0.196  &0.099  &0.065  &0.153  &0.118  &0.054  &0.109  &0.160\\
    \hline
    \textit{Average}      &0.023  &0.331  &0.169  &0.116  &0.075  &0.164  &0.124  &0.038  &0.132  &0.196\\
    \hline
    \end{tabular}
  \end{center}
\end{table*}

\subsection{Results and Analyses}
\begin{figure}[htbp]
  \centering
  \includegraphics[width=\linewidth]{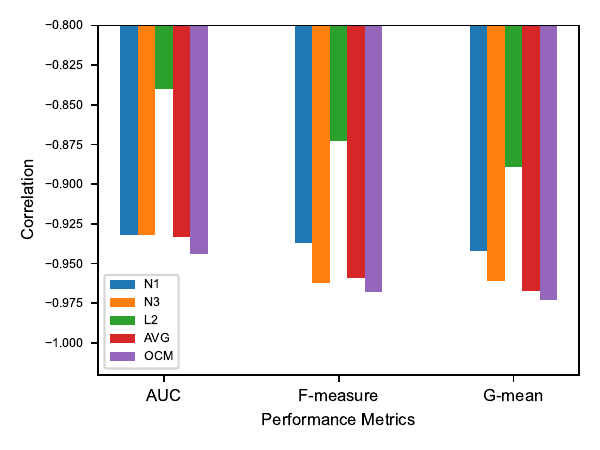}
  \caption{Correlation between data complexity measures and performance metrics.}
  \label{fig:correlation}
\end{figure}

First, we show the Pearson correlation between the data complexity measures and the performance metrics in Fig.~\ref{fig:correlation}. The performance results are obtained with the CART model. A highly statistically significant correlation is observed $(p<0.01)$. Among the performance metrics, AUC has a lower correlation than F-measure and G-mean, which show similar distributions. Among the individual complexity measures, L2 has the lowest correlation, probably because the base model is nonlinear, whereas N3 achieves the highest. In addition, we include the result of AVG, the naive average of the three measures. OCM shows a stronger correlation with all three performance metrics than either the individual measures or AVG, demonstrating the effectiveness of this overall complexity measure.

The average \textit{F-measure} of the compared methods on the Easy, Median, and Hard datasets is summarized in Table \ref{tab:f1}, and the results of the pairwise t-tests are shown in Table \ref{tab:test-f1}. The average \textit{G-mean} is summarized in Table \ref{tab:gmean}, with the corresponding t-test results in Table \ref{tab:test-gmean}.
In Tables \ref{tab:test-f1} and \ref{tab:test-gmean}, the three values at the intersection of row $i$ and column $j$ indicate, respectively, on how many datasets method $i$ is significantly better than, not significantly different from, or significantly worse than method $j$.

\begin{figure*}[htbp]
  \centering
  \includegraphics[width=0.8\textwidth]{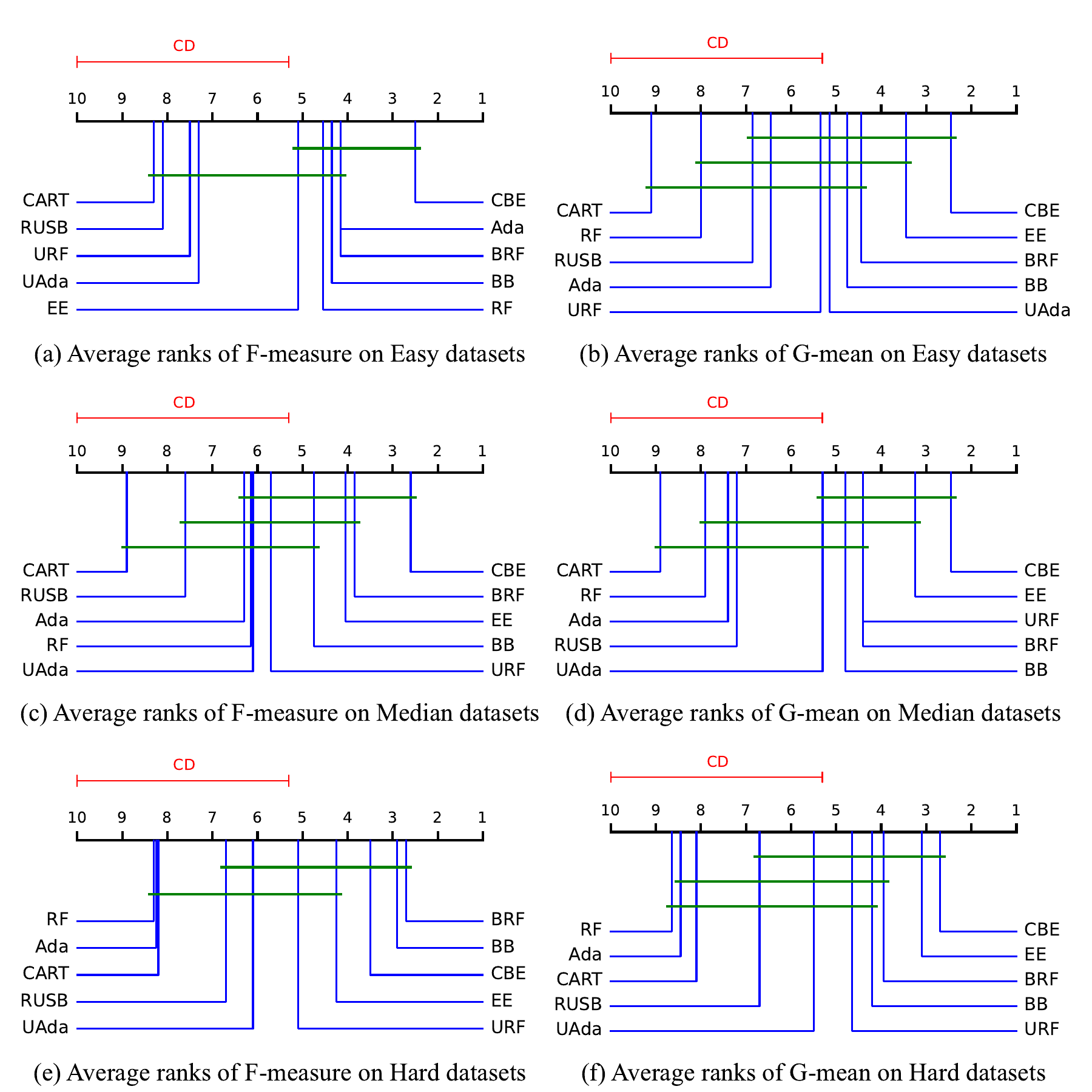}
  \caption{Average ranks of all methods and critical difference diagram on three kinds of datasets for the Nemenyi test. Groups of methods that are not significantly different at 0.05 are connected.}
  \label{fig:friedman-test}
\end{figure*}

Furthermore, we perform a Friedman test, which rejects the hypothesis that all algorithms are equivalent at a significance level of 0.05 \cite{b28}, and we obtain the average ranks of all methods for the two metrics. We then conduct a post-hoc Nemenyi test to compare the average ranks pairwise. According to this test, two classifiers differ significantly if their average ranks differ by at least the critical difference (CD = 4.24). Fig.~\ref{fig:friedman-test} summarizes these comparisons. The top line is the axis along which the average rank $R_m$ of each method $m$ is plotted, from the highest ranks (worst methods) on the left to the lowest ranks (best methods) on the right. Groups of methods that are not statistically different are connected.

On Easy datasets, the traditional methods Ada and RF achieve high F-measure and G-mean; in particular, Ada surpasses most imbalance learning methods on F-measure. On Median and Hard datasets, however, both Ada and RF degrade severely on both metrics. These results show that imbalance learning methods mainly benefit Median and Hard datasets.

On Median and Hard datasets, even simple undersampling allows UAda and URF to substantially alleviate the effects of class imbalance and improve significantly over Ada and RF on both metrics. Well-designed methods such as BB, BRF, and EE outperform UAda and URF on both metrics. BRF draws a balanced bootstrap for every tree, which reduces variance more effectively than EE, and inherits the random feature subspace technique from RF, which further reduces variance compared with BB. Hence, BRF performs better than both BB and EE.

Benefiting from the boundary-aware mechanism, CBE achieves the highest average F-measure and G-mean on Easy and Median datasets. On Hard datasets, it also attains near-optimal performance. In the pairwise t-tests, CBE outperforms the other methods on most datasets, except that it loses to BRF on F-measure for Hard datasets (0-6-2). The post-hoc Nemenyi test gives the same picture: CBE achieves the best rank for all three categories and both metrics, again except for BRF on F-measure for Hard datasets. The key is not uniform improvement but adaptive, complexity-aware sampling intensity. Guided by OCM, the method strengthens boundary sampling on easy and median datasets where a separable boundary exists, yielding consistent gains, and automatically relaxes it on hard datasets where intrinsic overlap bounds the Bayes error. It therefore delivers measurable gains without degrading performance on hard cases, an advantage over aggressive boundary sampling.

\begin{figure}[htbp]
  \centering
  \includegraphics[width=\linewidth]{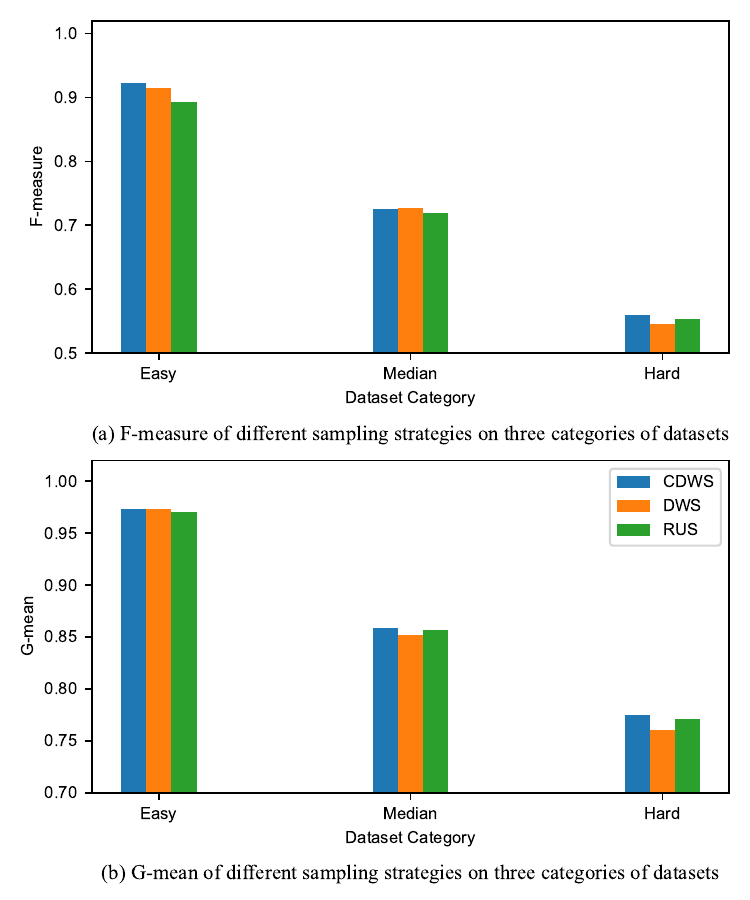}
  \caption{F-measure and G-mean of different sampling strategies on the three categories of datasets.}
  \label{fig:three-sample}
\end{figure}

We also conduct an experiment to examine the impact of different sampling strategies; the results are shown in Fig.~\ref{fig:three-sample}, where RUS denotes random under-sampling. On Easy datasets, CDWS performs best and RUS worst in terms of F-measure, confirming the effectiveness of the boundary-aware mechanism. CDWS outperforms DWS because it uses the $e^{-x}$ function to calculate the sampling weights, rather than the simple $1/x$ form, which tends to inflate the weights of noisy samples. On Median datasets, the three sampling methods perform similarly. On Hard datasets, CDWS and RUS achieve similar results owing to the complexity-adaptive design of boundary sampling intensity, and both outperform the aggressive boundary sampling of DWS. Similar results are observed for G-mean.

\begin{table*}
  \caption{F-measure of Compared Methods with Standard Deviation on Easy, Median, and Hard Datasets.}
  \label{tab:f1}
  \begin{center}
    \begin{tabular}{c|ccccccccc}
    \hline
    \textbf{Easy}        &mfeat-fac       &mfeat-pix      &mfeat-kar      &newthyroid2    &vowel0 &vehicle        &musk2             &ecoli4    &Average\\
    \hline
    CART      &0.914$\pm$0.032  &0.854$\pm$0.040  &0.764$\pm$0.056  &0.849$\pm$0.103  &0.881$\pm$0.081  &0.846$\pm$0.049  &0.886$\pm$0.019  &0.700$\pm$0.202  &0.8370\\
    Ada       &0.963$\pm$0.024  &0.950$\pm$0.021  &0.843$\pm$0.047  &0.924$\pm$0.085  &0.984$\pm$0.025  &0.949$\pm$0.021  &0.907$\pm$0.018  &0.729$\pm$0.252  &0.9061\\
    RF        &0.955$\pm$0.022  &0.941$\pm$0.022  &0.843$\pm$0.050  &0.945$\pm$0.063  &0.949$\pm$0.039  &0.927$\pm$0.029  &0.916$\pm$0.014  &0.677$\pm$0.288  &0.8942\\
    UAda      &0.886$\pm$0.077  &0.944$\pm$0.033  &0.928$\pm$0.069  &0.843$\pm$0.136  &0.802$\pm$0.101  &0.901$\pm$0.037  &0.871$\pm$0.021  &0.519$\pm$0.208  &0.8367\\
    URF       &0.905$\pm$0.045  &0.926$\pm$0.036  &0.923$\pm$0.035  &0.927$\pm$0.073  &0.801$\pm$0.065  &0.875$\pm$0.041  &0.808$\pm$0.030  &0.603$\pm$0.191  &0.8459\\
    BB        &0.935$\pm$0.033  &0.954$\pm$0.025  &0.942$\pm$0.023  &0.957$\pm$0.061  &0.871$\pm$0.049  &0.900$\pm$0.034  &0.896$\pm$0.017  &0.742$\pm$0.172  &0.8996\\
    BRF       &0.934$\pm$0.031  &0.956$\pm$0.023  &0.942$\pm$0.024  &0.961$\pm$0.071  &0.868$\pm$0.049  &0.902$\pm$0.036  &0.895$\pm$0.017  &0.752$\pm$0.142  &0.9012\\
    RUSB      &0.872$\pm$0.041  &0.907$\pm$0.045  &0.826$\pm$0.062  &0.908$\pm$0.100  &0.817$\pm$0.081  &0.880$\pm$0.035  &0.843$\pm$0.022  &0.590$\pm$0.192  &0.8304\\
    EE        &0.930$\pm$0.033  &0.952$\pm$0.025  &0.950$\pm$0.027  &0.914$\pm$0.083  &0.863$\pm$0.053  &0.906$\pm$0.038  &0.903$\pm$0.016  &0.669$\pm$0.183  &0.8857\\
    CBE       &0.973$\pm$0.020  &0.971$\pm$0.019  &0.963$\pm$0.021  &0.972$\pm$0.052  &0.904$\pm$0.055  &0.944$\pm$0.027  &0.943$\pm$0.011  &0.738$\pm$0.189  &\textbf{0.9260}\\
    \hline
    \hline
    \textbf{Median}        &wave           &ecoli1         &glass0         &ecoli3         &glass1         &pima           &bank           &vehicle1               &Average\\
    \hline
    CART      &0.755$\pm$0.016  &0.768$\pm$0.077  &0.681$\pm$0.098  &0.519$\pm$0.173  &0.618$\pm$0.094  &0.568$\pm$0.059  &0.570$\pm$0.024  &0.503$\pm$0.070  &0.6227\\
    Ada       &0.858$\pm$0.015  &0.758$\pm$0.104  &0.793$\pm$0.070  &0.558$\pm$0.183  &0.760$\pm$0.081  &0.614$\pm$0.053  &0.611$\pm$0.020  &0.531$\pm$0.057  &0.6853\\
    RF        &0.852$\pm$0.014  &0.775$\pm$0.094  &0.774$\pm$0.081  &0.595$\pm$0.186  &0.707$\pm$0.086  &0.634$\pm$0.049  &0.626$\pm$0.018  &0.511$\pm$0.062  &0.6842\\
    UAda      &0.843$\pm$0.013  &0.763$\pm$0.072  &0.777$\pm$0.073  &0.541$\pm$0.113  &0.756$\pm$0.057  &0.648$\pm$0.059  &0.673$\pm$0.018  &0.641$\pm$0.064  &0.7052\\
    URF       &0.847$\pm$0.013  &0.782$\pm$0.075  &0.763$\pm$0.076  &0.543$\pm$0.107  &0.706$\pm$0.078  &0.679$\pm$0.050  &0.678$\pm$0.018  &0.632$\pm$0.061  &0.7037\\
    BB        &0.853$\pm$0.014  &0.770$\pm$0.081  &0.800$\pm$0.076  &0.571$\pm$0.121  &0.730$\pm$0.072  &0.672$\pm$0.045  &0.691$\pm$0.017  &0.628$\pm$0.053  &0.7143\\
    BRF       &0.853$\pm$0.014  &0.776$\pm$0.083  &0.801$\pm$0.065  &0.574$\pm$0.122  &0.738$\pm$0.069  &0.673$\pm$0.048  &0.690$\pm$0.017  &0.634$\pm$0.056  &0.7172\\
    RUSB      &0.830$\pm$0.014  &0.760$\pm$0.076  &0.756$\pm$0.084  &0.532$\pm$0.160  &0.734$\pm$0.085  &0.627$\pm$0.054  &0.662$\pm$0.018  &0.601$\pm$0.064  &0.6878\\
    EE        &0.849$\pm$0.012  &0.775$\pm$0.074  &0.808$\pm$0.056  &0.578$\pm$0.110  &0.766$\pm$0.071  &0.664$\pm$0.052  &0.683$\pm$0.016  &0.653$\pm$0.055  &0.7218\\
    CBE       &0.862$\pm$0.014  &0.794$\pm$0.078  &0.831$\pm$0.052  &0.593$\pm$0.116  &0.770$\pm$0.071  &0.666$\pm$0.055  &0.687$\pm$0.017  &0.656$\pm$0.059  &\textbf{0.7324}\\
    \hline
    \hline
    \textbf{Hard}       &mfeat-fou      &vehicle3       &wine   &yeast1 &mfeat-mor      &haberman       &mfeat-zer      &abalone        &Average\\
    \hline
    CART      &0.462$\pm$0.066  &0.511$\pm$0.067  &0.499$\pm$0.031  &0.489$\pm$0.040  &0.303$\pm$0.056  &0.319$\pm$0.124  &0.196$\pm$0.057  &0.185$\pm$0.036  &0.3705\\
    Ada       &0.177$\pm$0.073  &0.482$\pm$0.066  &0.581$\pm$0.029  &0.519$\pm$0.041  &0.301$\pm$0.077  &0.295$\pm$0.085  &0.125$\pm$0.049  &0.076$\pm$0.030  &0.3194\\
    RF        &0.306$\pm$0.069  &0.486$\pm$0.056  &0.517$\pm$0.025  &0.546$\pm$0.044  &0.316$\pm$0.065  &0.310$\pm$0.117  &0.078$\pm$0.047  &0.039$\pm$0.024  &0.3248\\
    UAda      &0.613$\pm$0.042  &0.609$\pm$0.055  &0.547$\pm$0.019  &0.578$\pm$0.035  &0.616$\pm$0.048  &0.415$\pm$0.084  &0.514$\pm$0.051  &0.287$\pm$0.027  &0.5226\\
    URF       &0.626$\pm$0.043  &0.616$\pm$0.051  &0.537$\pm$0.021  &0.598$\pm$0.042  &0.652$\pm$0.047  &0.478$\pm$0.075  &0.529$\pm$0.046  &0.297$\pm$0.024  &0.5417\\
    BB        &0.642$\pm$0.041  &0.625$\pm$0.052  &0.584$\pm$0.023  &0.609$\pm$0.038  &0.655$\pm$0.045  &0.463$\pm$0.079  &0.560$\pm$0.043  &0.313$\pm$0.028  &0.5564\\
    BRF       &0.641$\pm$0.042  &0.628$\pm$0.048  &0.582$\pm$0.024  &0.611$\pm$0.041  &0.655$\pm$0.044  &0.465$\pm$0.075  &0.563$\pm$0.045  &0.313$\pm$0.030  &\textbf{0.5573}\\
    RUSB      &0.535$\pm$0.049  &0.589$\pm$0.056  &0.565$\pm$0.023  &0.551$\pm$0.043  &0.601$\pm$0.048  &0.402$\pm$0.083  &0.441$\pm$0.047  &0.282$\pm$0.026  &0.4958\\
    EE        &0.634$\pm$0.040  &0.630$\pm$0.049  &0.569$\pm$0.023  &0.603$\pm$0.036  &0.654$\pm$0.044  &0.443$\pm$0.079  &0.558$\pm$0.042  &0.299$\pm$0.023  &0.5485\\
    CBE       &0.640$\pm$0.036  &0.636$\pm$0.046  &0.581$\pm$0.022  &0.601$\pm$0.039  &0.654$\pm$0.043  &0.458$\pm$0.085  &0.567$\pm$0.044  &0.300$\pm$0.024  &0.5547\\
    \hline
    \end{tabular}
  \end{center}
\end{table*}

\begin{table*}
  \caption{Comparison of Pairwise t-tests on F-measure with Significance Level 0.05 on Easy, Median, and Hard Datasets.}
  \label{tab:test-f1}
  \begin{center}
  \begin{tabular}{c|cccccccccc}
    \hline
    F-measure &CART   & Ada     & RF     & UAda   & URF    & BB   & BRF     & RUSB    & EE   &CBE \\
    \hline
    CART      & —  & 0-1-7  & 0-1-7  & 2-2-4  & 2-1-5  & 0-1-7  & 0-1-7  & 0-4-4  & 0-2-6  & 0-0-8\\
    Ada   & 7-1-0      & —  & 4-2-2  & 6-1-1  & 6-1-1  & 4-2-2  & 4-2-2  & 6-2-0  & 3-4-1  & 1-2-5\\
    RF    & 7-1-0  & 2-2-4      & —  & 6-1-1  & 5-2-1  & 4-2-2  & 4-2-2  & 7-1-0  & 5-1-2  & 1-1-6\\
    UAda  & 4-2-2  & 1-1-6  & 1-1-6      & —  & 3-2-3  & 0-2-6  & 0-2-6  & 4-2-2  & 0-1-7  & 0-0-8\\
    URF   & 5-1-2  & 1-1-6  & 1-2-5  & 3-2-3      & —  & 0-0-8  & 0-0-8  & 3-4-1  & 0-1-7  & 0-0-8\\
    BB    & 7-1-0  & 2-2-4  & 2-2-4  & 6-2-0  & 8-0-0      & —  & 0-8-0  & 8-0-0  & 2-4-2  & 0-2-6\\
    BRF   & 7-1-0  & 2-2-4  & 2-2-4  & 6-2-0  & 8-0-0  & 0-8-0      & —  & 8-0-0  & 2-4-2  & 0-2-6\\
    RUSB  & 4-4-0  & 0-2-6  & 0-1-7  & 2-2-4  & 1-4-3  & 0-0-8  & 0-0-8      & —  & 0-1-7  & 0-0-8\\
    EE    & 6-2-0  & 1-4-3  & 2-1-5  & 7-1-0  & 7-1-0  & 2-4-2  & 2-4-2  & 7-1-0      & —  & 0-0-8\\
    CBE   & 8-0-0  & 5-2-1  & 6-1-1  & 8-0-0  & 8-0-0  & 6-2-0  & 6-2-0  & 8-0-0  & 8-0-0      & —\\
    \hline
    \hline
    CART      & —  & 0-1-7  & 0-2-6  & 0-1-7  & 0-1-7  & 0-1-7  & 0-1-7  & 0-1-7  & 0-1-7  & 0-1-7\\
    Ada   & 7-1-0      & —  & 3-3-2  & 1-4-3  & 3-2-3  & 2-3-3  & 2-3-3  & 2-3-3  & 1-4-3  & 0-3-5\\
    RF    & 6-2-0  & 2-3-3      & —  & 1-3-4  & 1-4-3  & 0-3-5  & 0-3-5  & 1-5-2  & 1-2-5  & 0-1-7\\
    UAda  & 7-1-0  & 3-4-1  & 4-3-1      & —  & 1-3-4  & 2-1-5  & 1-2-5  & 6-2-0  & 0-2-6  & 0-1-7\\
    URF   & 7-1-0  & 3-2-3  & 3-4-1  & 4-3-1      & —  & 1-2-5  & 0-3-5  & 5-2-1  & 1-2-5  & 1-0-7\\
    BB    & 7-1-0  & 3-3-2  & 5-3-0  & 5-1-2  & 5-2-1      & —  & 0-8-0  & 5-3-0  & 2-4-2  & 1-1-6\\
    BRF   & 7-1-0  & 3-3-2  & 5-3-0  & 5-2-1  & 5-3-0  & 0-8-0      & —  & 6-2-0  & 3-3-2  & 1-1-6\\
    RUSB  & 7-1-0  & 3-3-2  & 2-5-1  & 0-2-6  & 1-2-5  & 0-3-5  & 0-2-6      & —  & 0-1-7  & 0-0-8\\
    EE    & 7-1-0  & 3-4-1  & 5-2-1  & 6-2-0  & 5-2-1  & 2-4-2  & 2-3-3  & 7-1-0      & —  & 0-4-4\\
    CBE   & 7-1-0  & 5-3-0  & 7-1-0  & 7-1-0  & 7-0-1  & 6-1-1  & 6-1-1  & 8-0-0  & 4-4-0      & —\\
    \hline
    \hline
    CART      & —  & 3-3-2  & 3-3-2  & 0-0-8  & 0-0-8  & 0-0-8  & 0-0-8  & 0-0-8  & 0-0-8  & 0-0-8\\
    Ada   & 2-3-3      & —  & 3-3-2  & 1-0-7  & 1-0-7  & 0-1-7  & 0-1-7  & 1-0-7  & 1-0-7  & 0-1-7\\
    RF    & 2-3-3  & 2-3-3      & —  & 0-0-8  & 0-0-8  & 0-0-8  & 0-0-8  & 0-1-7  & 0-0-8  & 0-0-8\\
    UAda  & 8-0-0  & 7-0-1  & 8-0-0      & —  & 1-1-6  & 0-0-8  & 0-0-8  & 5-2-1  & 0-0-8  & 0-0-8\\
    URF   & 8-0-0  & 7-0-1  & 8-0-0  & 6-1-1      & —  & 0-2-6  & 0-2-6  & 7-0-1  & 1-3-4  & 1-3-4\\
    BB    & 8-0-0  & 7-1-0  & 8-0-0  & 8-0-0  & 6-2-0      & —  & 0-8-0  & 8-0-0  & 3-5-0  & 1-5-2\\
    BRF   & 8-0-0  & 7-1-0  & 8-0-0  & 8-0-0  & 6-2-0  & 0-8-0      & —  & 8-0-0  & 5-3-0  & 2-6-0\\
    RUSB  & 8-0-0  & 7-0-1  & 7-1-0  & 1-2-5  & 1-0-7  & 0-0-8  & 0-0-8      & —  & 0-1-7  & 0-0-8\\
    EE    & 8-0-0  & 7-0-1  & 8-0-0  & 8-0-0  & 4-3-1  & 0-5-3  & 0-3-5  & 7-1-0      & —  & 0-5-3\\
    CBE   & 8-0-0  & 7-1-0  & 8-0-0  & 8-0-0  & 4-3-1  & 2-5-1  & 0-6-2  & 8-0-0  & 3-5-0      & —\\
    \hline
    \end{tabular}
  \end{center}
\end{table*}

\begin{table*}
  \caption{G-mean of Compared Methods with Standard Deviation on Easy, Median, and Hard Datasets.}
  \label{tab:gmean}
  \begin{center}
    \begin{tabular}{c|cccccccccc}
    \hline
    \textbf{Easy}         &mfeat-fac              &mfeat-pix              &mfeat-kar              &newthyroid2            &vowel0         &vehicle                &musk2          &ecoli4         &Average\\
    \hline
    CART      &0.946$\pm$0.027  &0.914$\pm$0.032  &0.858$\pm$0.046  &0.915$\pm$0.071  &0.937$\pm$0.053  &0.896$\pm$0.039  &0.931$\pm$0.014  &0.834$\pm$0.178  &0.9040\\
    Ada       &0.967$\pm$0.021  &0.951$\pm$0.020  &0.855$\pm$0.042  &0.945$\pm$0.070  &0.985$\pm$0.024  &0.968$\pm$0.016  &0.913$\pm$0.017  &0.779$\pm$0.239  &0.9204\\
    RF        &0.958$\pm$0.021  &0.944$\pm$0.022  &0.855$\pm$0.044  &0.951$\pm$0.057  &0.958$\pm$0.035  &0.954$\pm$0.021  &0.924$\pm$0.013  &0.706$\pm$0.277  &0.9061\\
    UAda      &0.973$\pm$0.023  &0.984$\pm$0.011  &0.975$\pm$0.024  &0.938$\pm$0.055  &0.966$\pm$0.025  &0.964$\pm$0.014  &0.958$\pm$0.010  &0.869$\pm$0.160  &0.9535\\
    URF       &0.977$\pm$0.013  &0.973$\pm$0.013  &0.966$\pm$0.016  &0.980$\pm$0.036  &0.972$\pm$0.013  &0.954$\pm$0.016  &0.939$\pm$0.010  &0.931$\pm$0.071  &0.9613\\
    BB        &0.976$\pm$0.017  &0.973$\pm$0.016  &0.955$\pm$0.020  &0.987$\pm$0.033  &0.979$\pm$0.017  &0.962$\pm$0.014  &0.956$\pm$0.008  &0.932$\pm$0.091  &0.9651\\
    BRF       &0.976$\pm$0.015  &0.975$\pm$0.014  &0.955$\pm$0.019  &0.987$\pm$0.036  &0.978$\pm$0.017  &0.963$\pm$0.014  &0.956$\pm$0.008  &0.942$\pm$0.073  &0.9667\\
    RUSB      &0.967$\pm$0.018  &0.971$\pm$0.013  &0.950$\pm$0.023  &0.969$\pm$0.041  &0.966$\pm$0.022  &0.943$\pm$0.019  &0.939$\pm$0.011  &0.892$\pm$0.116  &0.9497\\
    EE        &0.985$\pm$0.012  &0.984$\pm$0.011  &0.974$\pm$0.015  &0.979$\pm$0.029  &0.979$\pm$0.015  &0.967$\pm$0.013  &0.966$\pm$0.008  &0.939$\pm$0.071  &0.9716\\
    CBE       &0.989$\pm$0.012  &0.984$\pm$0.012  &0.973$\pm$0.019  &0.987$\pm$0.039  &0.985$\pm$0.012  &0.976$\pm$0.012  &0.972$\pm$0.008  &0.945$\pm$0.080  &\textbf{0.9763}\\
    \hline
    \hline
    \textbf{Median}        &wave           &ecoli1         &glass0         &ecoli3         &glass1         &pima           &bank           &vehicle1               &Average\\
    \hline
    CART      &0.813$\pm$0.013  &0.852$\pm$0.060  &0.760$\pm$0.077  &0.668$\pm$0.173  &0.700$\pm$0.073  &0.660$\pm$0.047  &0.699$\pm$0.019  &0.642$\pm$0.051  &0.7241\\
    Ada       &0.896$\pm$0.011  &0.830$\pm$0.077  &0.839$\pm$0.062  &0.672$\pm$0.160  &0.805$\pm$0.067  &0.696$\pm$0.038  &0.705$\pm$0.018  &0.649$\pm$0.042  &0.7615\\
    RF        &0.891$\pm$0.010  &0.856$\pm$0.068  &0.826$\pm$0.063  &0.693$\pm$0.158  &0.760$\pm$0.071  &0.711$\pm$0.035  &0.718$\pm$0.016  &0.631$\pm$0.050  &0.7604\\
    UAda      &0.901$\pm$0.008  &0.882$\pm$0.047  &0.842$\pm$0.059  &0.865$\pm$0.062  &0.810$\pm$0.045  &0.726$\pm$0.039  &0.824$\pm$0.010  &0.782$\pm$0.036  &0.8291\\
    URF       &0.902$\pm$0.009  &0.900$\pm$0.044  &0.830$\pm$0.064  &0.877$\pm$0.051  &0.769$\pm$0.063  &0.752$\pm$0.032  &0.830$\pm$0.009  &0.775$\pm$0.036  &0.8293\\
    BB        &0.904$\pm$0.009  &0.888$\pm$0.049  &0.860$\pm$0.059  &0.872$\pm$0.072  &0.786$\pm$0.057  &0.747$\pm$0.027  &0.828$\pm$0.010  &0.766$\pm$0.032  &0.8313\\
    BRF       &0.904$\pm$0.009  &0.891$\pm$0.047  &0.862$\pm$0.051  &0.871$\pm$0.073  &0.793$\pm$0.055  &0.748$\pm$0.029  &0.827$\pm$0.010  &0.770$\pm$0.034  &0.8332\\
    RUSB      &0.886$\pm$0.010  &0.872$\pm$0.047  &0.824$\pm$0.064  &0.807$\pm$0.121  &0.790$\pm$0.068  &0.710$\pm$0.035  &0.800$\pm$0.012  &0.743$\pm$0.042  &0.8041\\
    EE        &0.904$\pm$0.008  &0.894$\pm$0.040  &0.865$\pm$0.049  &0.887$\pm$0.051  &0.819$\pm$0.050  &0.740$\pm$0.031  &0.833$\pm$0.008  &0.793$\pm$0.031  &0.8418\\
    CBE       &0.910$\pm$0.009  &0.899$\pm$0.043  &0.880$\pm$0.040  &0.887$\pm$0.058  &0.821$\pm$0.056  &0.741$\pm$0.035  &0.837$\pm$0.008  &0.794$\pm$0.033  &\textbf{0.8462}\\
    \hline
    \hline
    \textbf{Hard}       &mfeat-fou      &vehicle3       &wine   &yeast1 &mfeat-mor      &haberman       &mfeat-zer      &abalone        &Average\\
    \hline
    CART      &0.651$\pm$0.052  &0.652$\pm$0.055  &0.668$\pm$0.025  &0.621$\pm$0.032  &0.523$\pm$0.061  &0.482$\pm$0.118  &0.418$\pm$0.072  &0.400$\pm$0.046  &0.5517\\
    Ada       &0.312$\pm$0.077  &0.602$\pm$0.054  &0.663$\pm$0.022  &0.632$\pm$0.033  &0.523$\pm$0.079  &0.462$\pm$0.083  &0.304$\pm$0.081  &0.204$\pm$0.044  &0.4628\\
    RF        &0.447$\pm$0.065  &0.602$\pm$0.048  &0.612$\pm$0.020  &0.648$\pm$0.034  &0.535$\pm$0.068  &0.453$\pm$0.120  &0.226$\pm$0.093  &0.134$\pm$0.051  &0.4573\\
    UAda      &0.901$\pm$0.022  &0.761$\pm$0.040  &0.793$\pm$0.011  &0.704$\pm$0.024  &0.894$\pm$0.023  &0.574$\pm$0.065  &0.839$\pm$0.031  &0.636$\pm$0.024  &0.7628\\
    URF       &0.907$\pm$0.020  &0.766$\pm$0.033  &0.789$\pm$0.013  &0.721$\pm$0.027  &0.928$\pm$0.015  &0.635$\pm$0.061  &0.839$\pm$0.029  &0.647$\pm$0.021  &0.7790\\
    BB        &0.901$\pm$0.024  &0.767$\pm$0.038  &0.799$\pm$0.015  &0.726$\pm$0.025  &0.932$\pm$0.013  &0.622$\pm$0.070  &0.840$\pm$0.028  &0.634$\pm$0.025  &0.7776\\
    BRF       &0.899$\pm$0.024  &0.769$\pm$0.035  &0.797$\pm$0.015  &0.728$\pm$0.028  &0.932$\pm$0.013  &0.624$\pm$0.068  &0.842$\pm$0.030  &0.635$\pm$0.027  &0.7783\\
    RUSB      &0.846$\pm$0.034  &0.736$\pm$0.043  &0.790$\pm$0.016  &0.676$\pm$0.030  &0.876$\pm$0.031  &0.568$\pm$0.068  &0.782$\pm$0.034  &0.607$\pm$0.029  &0.7351\\
    EE        &0.912$\pm$0.017  &0.779$\pm$0.036  &0.808$\pm$0.013  &0.725$\pm$0.026  &0.933$\pm$0.012  &0.600$\pm$0.061  &0.867$\pm$0.022  &0.651$\pm$0.022  &0.7844\\
    CBE       &0.915$\pm$0.017  &0.783$\pm$0.030  &0.813$\pm$0.013  &0.724$\pm$0.025  &0.933$\pm$0.011  &0.615$\pm$0.066  &0.868$\pm$0.025  &0.651$\pm$0.020  &\textbf{0.7876}\\
    \hline
    \end{tabular}
  \end{center}
\end{table*}

\begin{table*}
  \caption{Comparison of Pairwise t-tests on G-mean with Significance Level 0.05 on Easy, Median, and Hard Datasets.}
  \label{tab:test-gmean}
  \begin{center}
  \begin{tabular}{c|cccccccccc}
    \hline
    G-mean    &CART   & Ada     & RF     & UAda   & URF    & BB   & BRF     & RUSB    & EE   &CBE \\
    \hline
    CART      & —  & 0-1-7  & 0-1-7  & 0-0-8  & 0-0-8  & 0-0-8  & 0-0-8  & 0-0-8  & 0-0-8  & 0-0-8\\
    Ada   & 7-1-0      & —  & 5-2-1  & 1-3-4  & 2-0-6  & 1-1-6  & 1-1-6  & 2-1-5  & 1-1-6  & 0-1-7\\
    RF    & 7-1-0  & 1-2-5      & —  & 0-2-6  & 0-1-7  & 0-0-8  & 0-0-8  & 1-1-6  & 0-0-8  & 0-0-8\\
    UAda  & 8-0-0  & 4-3-1  & 6-2-0      & —  & 4-2-2  & 2-3-3  & 2-3-3  & 4-3-1  & 0-3-5  & 0-2-6\\
    URF   & 8-0-0  & 6-0-2  & 7-1-0  & 2-2-4      & —  & 1-3-4  & 1-3-4  & 5-3-0  & 0-2-6  & 0-2-6\\
    BB    & 8-0-0  & 6-1-1  & 8-0-0  & 3-3-2  & 4-3-1      & —  & 0-8-0  & 6-2-0  & 0-3-5  & 0-2-6\\
    BRF   & 8-0-0  & 6-1-1  & 8-0-0  & 3-3-2  & 4-3-1  & 0-8-0      & —  & 6-2-0  & 0-3-5  & 0-2-6\\
    RUSB  & 8-0-0  & 5-1-2  & 6-1-1  & 1-3-4  & 0-3-5  & 0-2-6  & 0-2-6      & —  & 0-1-7  & 0-0-8\\
    EE    & 8-0-0  & 6-1-1  & 8-0-0  & 5-3-0  & 6-2-0  & 5-3-0  & 5-3-0  & 7-1-0      & —  & 0-4-4\\
    CBE   & 8-0-0  & 7-1-0  & 8-0-0  & 6-2-0  & 6-2-0  & 6-2-0  & 6-2-0  & 8-0-0  & 4-4-0      & —\\
    \hline
    \hline
    CART      & —  & 0-2-6  & 0-2-6  & 0-0-8  & 0-0-8  & 0-0-8  & 0-0-8  & 0-0-8  & 0-0-8  & 0-0-8\\
    Ada   & 6-2-0      & —  & 3-2-3  & 0-2-6  & 1-1-6  & 1-0-7  & 1-0-7  & 1-2-5  & 0-1-7  & 0-0-8\\
    RF    & 6-2-0  & 3-2-3      & —  & 0-1-7  & 0-2-6  & 0-0-8  & 0-0-8  & 1-2-5  & 0-0-8  & 0-0-8\\
    UAda  & 8-0-0  & 6-2-0  & 7-1-0      & —  & 1-4-3  & 2-2-4  & 2-3-3  & 7-1-0  & 0-1-7  & 0-1-7\\
    URF   & 8-0-0  & 6-1-1  & 6-2-0  & 3-4-1      & —  & 1-5-2  & 2-4-2  & 6-2-0  & 1-2-5  & 1-2-5\\
    BB    & 8-0-0  & 7-0-1  & 8-0-0  & 4-2-2  & 2-5-1      & —  & 0-8-0  & 7-1-0  & 1-3-4  & 0-1-7\\
    BRF   & 8-0-0  & 7-0-1  & 8-0-0  & 3-3-2  & 2-4-2  & 0-8-0      & —  & 7-1-0  & 1-3-4  & 0-2-6\\
    RUSB  & 8-0-0  & 5-2-1  & 5-2-1  & 0-1-7  & 0-2-6  & 0-1-7  & 0-1-7      & —  & 0-0-8  & 0-0-8\\
    EE    & 8-0-0  & 7-1-0  & 8-0-0  & 7-1-0  & 5-2-1  & 4-3-1  & 4-3-1  & 8-0-0      & —  & 0-5-3\\
    CBE   & 8-0-0  & 8-0-0  & 8-0-0  & 7-1-0  & 5-2-1  & 7-1-0  & 6-2-0  & 8-0-0  & 3-5-0      & —\\
    \hline
    \hline
    CART      & —  & 4-2-2  & 5-2-1  & 0-0-8  & 0-0-8  & 0-0-8  & 0-0-8  & 0-0-8  & 0-0-8  & 0-0-8\\
    Ada   & 2-2-4      & —  & 3-3-2  & 0-0-8  & 0-0-8  & 0-0-8  & 0-0-8  & 0-0-8  & 0-0-8  & 0-0-8\\
    RF    & 1-2-5  & 2-3-3      & —  & 0-0-8  & 0-0-8  & 0-0-8  & 0-0-8  & 0-0-8  & 0-0-8  & 0-0-8\\
    UAda  & 8-0-0  & 8-0-0  & 8-0-0      & —  & 1-2-5  & 0-4-4  & 0-4-4  & 6-2-0  & 0-0-8  & 0-0-8\\
    URF   & 8-0-0  & 8-0-0  & 8-0-0  & 5-2-1      & —  & 2-4-2  & 2-3-3  & 7-1-0  & 1-3-4  & 1-2-5\\
    BB    & 8-0-0  & 8-0-0  & 8-0-0  & 4-4-0  & 2-4-2      & —  & 0-8-0  & 8-0-0  & 1-2-5  & 0-3-5\\
    BRF   & 8-0-0  & 8-0-0  & 8-0-0  & 4-4-0  & 3-3-2  & 0-8-0      & —  & 8-0-0  & 1-2-5  & 0-3-5\\
    RUSB  & 8-0-0  & 8-0-0  & 8-0-0  & 0-2-6  & 0-1-7  & 0-0-8  & 0-0-8      & —  & 0-0-8  & 0-0-8\\
    EE    & 8-0-0  & 8-0-0  & 8-0-0  & 8-0-0  & 4-3-1  & 5-2-1  & 5-2-1  & 8-0-0      & —  & 0-7-1\\
    CBE   & 8-0-0  & 8-0-0  & 8-0-0  & 8-0-0  & 5-2-1  & 5-3-0  & 5-3-0  & 8-0-0  & 1-7-0      & —\\
    \hline
    \end{tabular}
  \end{center}
\end{table*}

\section{Conclusion} \label{sec:conclusion}
In this paper, we propose a complexity-adaptive boundary-aware ensemble framework, \texttt{CBE}, for class imbalance classification. The framework consists of three components: an overall complexity measure OCM, a novel distance-weighted under-sampling method CDWS, and a simple ensemble strategy adapted from EasyEnsemble. Experiments on 24 benchmark datasets demonstrate the effectiveness of OCM and the superiority of CBE over several typical ensemble methods in terms of F-measure and G-mean, with a particularly pronounced advantage in F-measure. We also examine the impact of different sampling strategies. Guided by OCM, CBE strengthens boundary sampling on easy and median datasets where a separable boundary exists and automatically relaxes it on hard datasets where intrinsic overlap bounds the Bayes error.

Introducing complexity measures to adjust the sampling intensity of boundary examples also incurs computational overhead, because computing the data complexity and pairwise Euclidean distances scales quadratically with the dataset size and dimension. This may limit the applicability of the method to large-scale datasets and real-time scenarios.

\bibliographystyle{IEEEtran}
\bibliography{reference}

\end{document}